\documentclass[sigconf, screen]{acmart}

\AtBeginDocument{%
	\providecommand\BibTeX{{%
			\normalfont B\kern-0.5em{\scshape i\kern-0.25em b}\kern-0.8em\TeX}}}

\renewcommand\footnotetextcopyrightpermission[1]{} 

\usepackage{multirow}
\usepackage{wasysym}
\usepackage{makecell}
\usepackage{enumitem}
\begin{document}
	
	\title{A Robust Incomplete Multimodal Low-Rank Adaptation  Approach for  Emotion Recognition}
	
	 \author{Xinkui Zhao, Jinsong Shu, Guanjie Cheng, Zihe Liu, Naibo Wang, Shuiguang Deng, Zhongle Xie, Jianwei Yin, Yangyang Wu}
	 \authornote{Corresponding author: Yangyang Wu.} 
	 \affiliation{%
		   \institution{Zhejiang University}
		   \city{Zhejiang}
		   \country{China}
		 }
	
	
	
	\begin{abstract}
Multimodal Emotion Recognition (MER) often encounters incomplete multimodality in practical applications due to sensor failures or privacy protection requirements. While existing methods attempt to address various incomplete multimodal scenarios by balancing the training of each modality combination through additional gradients, these approaches face a critical limitation: training gradients from different modality combinations conflict with each other, ultimately degrading the performance of the final prediction model.
In this paper, we propose a \emph{unimodal decoupled dynamic low-rank adaptation} method based on modality combinations, named \textsf{MCULoRA}, which is a novel framework for the parameter-efficient training of incomplete multimodal learning models.
\textsf{MCULoRA} consists of two key modules, \emph{modality combination aware low-rank adaptation} (MCLA) and \emph{dynamic parameter fine-tuning} (DPFT).
The MCLA module effectively decouples the shared information from the distinct characteristics of individual modality combinations.
The DPFT module adjusts the training ratio of modality combinations based on the separability of each modality's representation space, optimizing the learning efficiency across different modality combinations.
Our extensive experimental evaluation in multiple benchmark datasets demonstrates that \textsf{MCULoRA} substantially outperforms previous incomplete multimodal learning approaches in downstream task accuracy.

	\end{abstract}

\ccsdesc[300]{Information systems~Multimedia information systems}
\ccsdesc[500]{Information systems~Sentiment analysis}
\ccsdesc[300]{Information systems~Multimedia and multimodal retrieval}
	
	\keywords{Incomplete multimodal learning, Low-rank adaptation}
	
	
	
	\maketitle
	
	\section{Introduction}
	
	Multimodal learning, which mines complementary information across modalities \cite{baltrusaitis2019multimodal,xu2023multimodal}, has advanced automatic emotion recognition significantly \cite{li2023decoupled,tsai2019multimodal,zhao2021emotion}. Multimodal Emotion Recognition (MER) has been applied in human-computer interaction \cite{cowie2001emotion,kirchner2019embedded}, dialogue systems \cite{fu2022learning,liang2022emotional}, and social media analysis \cite{somandepalli2021computational}.
    
    While existing MER methods achieve excellent performance with complete modal information \cite{li2023decoupled,gao2024embracing}, real-world scenarios frequently present incomplete multimodal data due to sensor failures, speech recognition errors, or privacy constraints \cite{zhao2021missing}. Models trained on complete data suffer significant performance degradation when processing incomplete inputs \cite{aguilar2019multimodal,parthasarathy2020training,pham2019found}, necessitating efficient adaptation approaches. Low-rank adaptation techniques offer a promising solution by efficiently adjusting model parameters to handle missing modalities without sacrificing computational efficiency, highlighting the practical importance of developing effective incomplete multimodal learning methods for emotion recognition.

    \begin{figure}[!t] 
		\centering
		 \includegraphics[width=1.0\columnwidth]{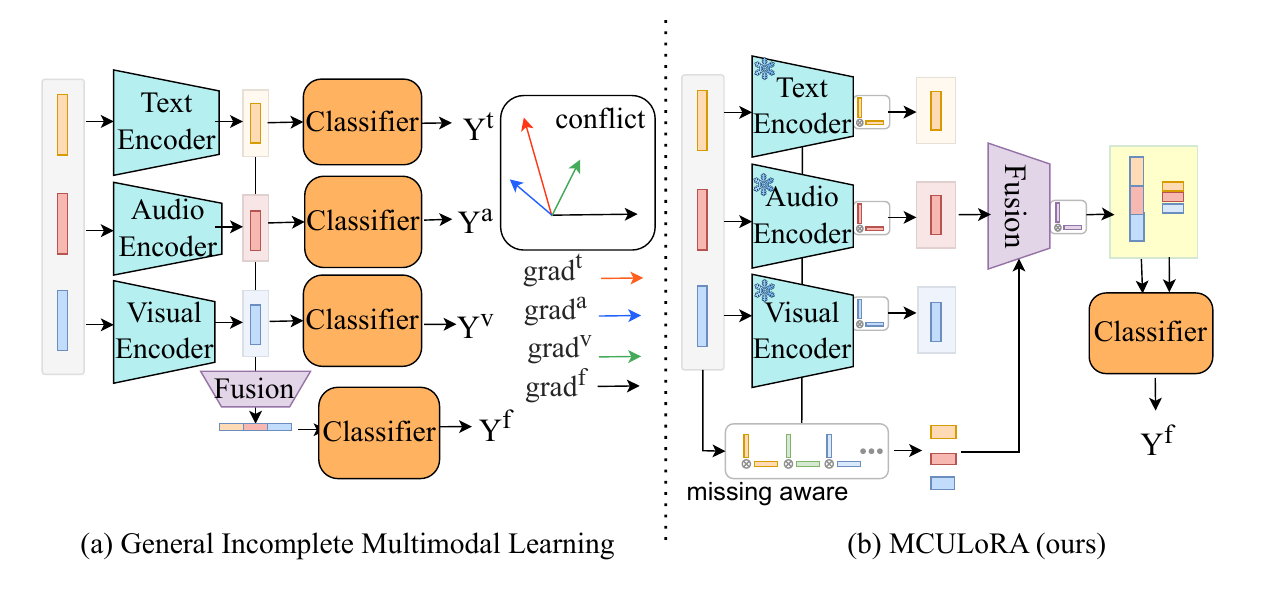} 
		\caption{(a) Existing incomplete multimodal learning methods add unimodal prediction losses to enhance characteristic information in fused representations, but suffer from persistent gradient conflicts between modality combinations. (b) Our MCULoRA approach with the MCLA module effectively decouples characteristic and common information in unimodal data, addressing the deficiency of characteristic information in joint multimodal representations.}
		\label{fig:settings}
	\end{figure}

    Approaches to incomplete multimodality generally fall into two categories: (1) Modality imputation techniques using generative models like VAEs \cite{shi2019variational,wu2018multimodal}, GANs \cite{cai2018deep}, and diffusion models \cite{wang2023incomplete} can effectively restore missing modalities but incur high computational costs that prohibit real-time applications; and (2) Joint representation learning methods \cite{pham2019found,zhao2021missing} that create consistent cross-modal representations but often sacrifice characteristic unimodal information. Recent advances attempt to address this limitation by extracting unimodal characteristic information through multi-stage learning or uncertainty-based approaches \cite{gao2024enhanced,xu2024leveraging}. Nevertheless, conflicts persist in the requisites for characteristic information across diverse modal combinations.

    Although existing advanced methods often add additional modality combination prediction losses to enhance the feature information after multi-modal fusion, as shown in Fig. \ref{fig:settings} (a). However, due to the different requirements of each modality combination for specific modality information, gradient conflicts are likely to occur between the modality combination losses, ultimately hindering the model from achieving the best performance. To achieve the best performance, traditional methods have to train independent models for each missing modality combination. In the inference stage, the corresponding model must also be selected according to the specific missing modality combination, which leads to an exponential increase in model parameters and training time during training as the number of modalities increases.
    
    Our research significantly differs from traditional incomplete multimodal learning methods. We have discovered that the characteristic information provided by single modalities varies across different modality combinations. Thus, the core challenge in efficiently fine-tuning incomplete multimodal models lies in how to fully explore the unique feature information required for each modality combination after the completion of single-modal representation.
    Based on this, we propose the \textsf{MCULoRA} framework, a brand-new architecture specifically designed for the efficient parameter fine-tuning of incomplete multimodal models. \textsf{MCULoRA} successfully decouples the parameter space by strategically combining low-rank decomposition modules related to and independent of modality combinations, enabling the model to effectively distinguish between the learning of modal common information and the learning of feature discriminative information for various modality combinations. At the same time, taking into account the differences in the abilities of different modality combinations to extract unimodal feature information, we have designed a mechanism that can dynamically adjust the overall probability of the occurrence of modality combinations according to the difficulty of decoupling the current unimodal representation space, so as to balance the learning process of unimodalities among different modality combinations. 
   Our contributions can be summarized as follows:
   \begin{itemize}
        \item We identified the flaws of traditional joint representation learning in incomplete multimodal scenarios. In response to these flaws, we proposed \textsf{MCULoRA}, a parameter-efficient training method that uses the characteristic information in unimodal data to assist the consistent multimodal fusion representation in making predictions. 
        \item We observed that there are differences in the extraction of characteristic information from unimodal representations under different combinations. Therefore, we designed a dynamic parameter fine-tuning strategy to balance the learning process of unimodal data in different modality combinations, thus enhancing the adaptability of the model.
        \item Through extensive experiments on public real-world multimodal emotion recognition datasets, MCULoRA demonstrated superior performance across various modality missing patterns, outperforming previous state-of-the-art methods with significant average accuracy improvements of 2.34\% and 6.04\% respectively.
    \end{itemize}
	
	\section{Related Work}
	
    \textbf{Incomplete multimodal learning} in MER is crucial for real-world applications where modal data may be missing. A direct means to address the problem of incomplete multimodality is data imputation. Unsupervised imputation methods include zero-value imputation and mean-value imputation \cite{ma2021maximum,parthasarathy2020training,zhang2022deep,john2023progressive,shi2024passion}.
    In recent years, deep learning based methods have used the missing modalities as supervisory information and employed generative models such as variational autoencoders \cite{shi2019variational,wu2018multimodal} and generative adversarial networks \cite{cai2018deep} to generate these missing modalities. Tran et al. \cite{tran2017missing} proposed the cascaded residual autoencoder, which simulates the correlations between different modalities by stacking residual autoencoders and then imputes the missing modalities based on the available ones. Wang et al. \cite{wang2023incomplete} used a score-based diffusion model to map random noise into the distribution space of the missing modalities and used the available modalities as semantic conditions to guide the denoising process, thereby restoring the missing modalities.
    
    However, imputation methods' high computational cost limits their real-time application. Consequently, approaches learning cross-modal joint representations through consistency constraints have gained prominence in incomplete multimodal emotion recognition. These include the Multimodal Cyclic Translation Network (MCTN) \cite{pham2019found}, which employs cyclic translation loss; Missing Modality Imagination Network (MMIN) \cite{zhao2021missing}, which uses cross-modal imagination with cyclic consistency; IF-MMIN \cite{zuo2023exploiting}, which extracts invariant features between modalities; Multimodal Reconstruction and Alignment Network (MRAN) \cite{luo2023multimodal}, which projects visual and audio features into text space; graph completion network (GCNet) \cite{lian2023gcnet}, which leverages graph neural networks for multimodal interaction. Nevertheless, consistency-based joint representations often neglect characteristic information of different modal combinations. To address this limitation, recent approaches employ mixture-of-experts (MoE) architectures, such as uncertainty-based routing mechanism \cite{gao2024enhanced} and two-stage hybrid MoE \cite{xu2024leveraging} that balances single-modal discrimination with cross-modal commonalities.
        
    Different from the above methods, the method we proposed introduces a modality combination aware low-rank adaptation module to decouple the demand for characteristic information from different modality combinations. Then, the unimodal shared adaptation module reduces the information redundancy among the characteristic information, enabling the  modality combination aware module to focus on the characteristic information. This allows the model to strike a balance between learning the common modal information and the characteristic information of modality combinations. This significantly enhances the discriminative ability of the representations used for inference. 
    	
    \textbf{Parameter-Efficient Training (PEFT)} is becoming increasingly important, especially when dealing with large-scale pre-trained models \cite{gao2023llama,hu2022lora,hu2023llm,zhang2023llama}. This is because traditional fine-tuning methods require adjusting a large number of parameters of the model for specific tasks, which may consume a vast amount of resources. In this field, two common techniques are adapters \cite{gao2023llama,zhang2023llama} and Low-Rank Adaptation (LoRA) \cite{dettmers2023qlora,hu2022lora}. Adapters are lightweight modules inserted between the layers of a pre-trained model. They allow for targeted modifications to the behavior of the model without changing the original pre-trained weights. This approach reduces the number of parameters that need to be fine-tuned, thereby decreasing the computational burden. Adapters have been proven effective in various tasks and provide a flexible and efficient way to adapt large models to specific tasks or datasets.
    
    However, one limitation of adapters is that they introduce additional parameters, which may increase the computational requirements during the inference process. On the other hand, LoRA offers a different approach to efficient parameter training. It uses low-rank decomposition to modify the weight matrices of the pre-trained model, enabling fine-tuning while maintaining the original structure and size of the weight matrices. The key advantage of LoRA is that it does not introduce additional parameters during the model's operation. Instead, it updates the pre-existing weights to enhance the model's performance on new tasks with minimal computational overhead. LoRA has been successfully applied in multiple fields, including Natural Language Processing \cite{chavan2023one,chen2023longlora,dettmers2023qlora,hu2022lora} and Computer Vision \cite{he2023parameter}, which demonstrates its versatility and effectiveness. However, in the face of incomplete multimodal scenarios, there is still no effective low-rank adaptation method to supplement the discriminative information of multimodal representations.

    \begin{figure*}[!t] 
        \centering
        \includegraphics[width=0.975\textwidth]{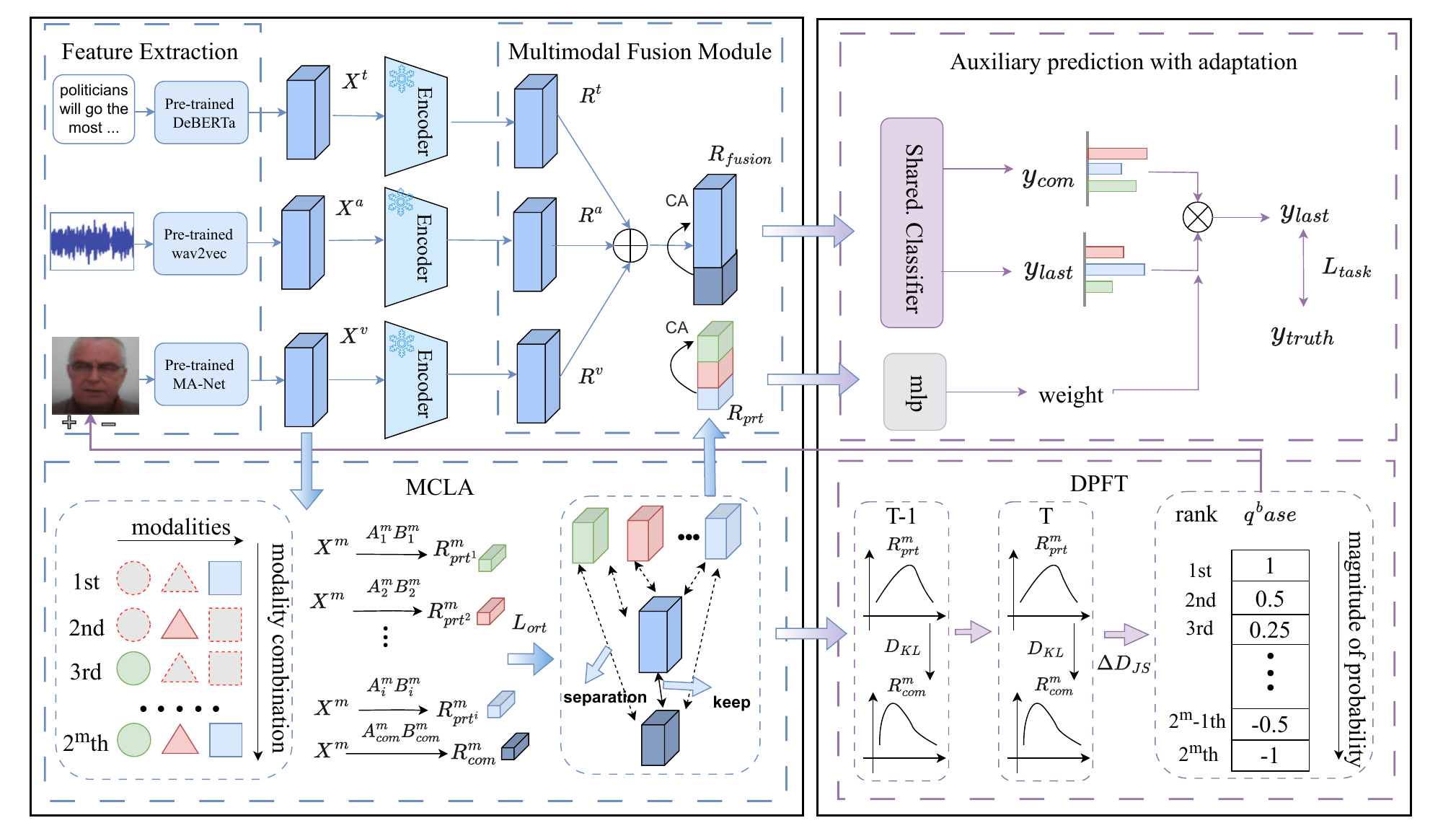}
        \caption{The overall framework of MCULoRA is as follows: During the training phase, in the first cell at the top, MCULoRA first extracts features from the original data. Subsequently, in the first cell at the bottom, the MCLA module decouples the unimodal representations. Finally, in the second cell at the top, the model leverages feature information from modality combination-aware adaptation to assist the joint representation in completing the prediction task. During training, the DPFT module in the second cell at the bottom dynamically adjusts the occurrence probability of different modality combinations based on the current decoupling status of individual modalities, thereby balancing the adaptation degree of single modalities across different combinations. Here, *m* denotes the maximum number of input modalities supported by the current model, and the CA operation refers to the cross-attention mechanism.}
        \vspace{0.03in}
        \label{fig:framework}
    \end{figure*}
    
	\section{Methodology}

	\subsection{Problem Definition}
	\label{sec:problem}
    
	In this subsection, we first examine the problem posed by incomplete multimodal learning in multimodal emotion recognition (MER), followed by our proposed approach to address this problem.
\vspace{0.02in}

    \textbf{Pre-define.} Referring to previous research \cite{lian2023gcnet}, we focus on incomplete multimodal learning in conversations, where three modalities, namely audio \(a\), text \(t\), and vision \(v\), are utilized. A conversation is denoted as \(G=\{(u_i,y_i)\}_{i = 1}^{L}\), where \(L\) is the number of utterances in the conversation, \(u_i\) is the \(i\)-th utterance in conversation \(G\), and \(y_i\) is the label corresponding to \(u_i\).
    \vspace{0.02in}
    
    When in the case of complete multimodality, \(X = \{X^m|m\in\{a,t,v\}\}\) forms the multimodal feature set, where \(X^m=\{x_i^m\}_{i = 1}^{L}\) is the single-modal feature set. In the case of incomplete multimodality, \(X^M=\{X^m|m\in M\}\) represents the feature set of available modalities, and \(\hat{X}^{\hat{M}}=\{\hat{X}^{\hat{m}}|\hat{m}\in\hat{M}\}\) represents the feature set of missing modalities, with \(M\cup\hat{M}=\{a,t,v\}\) and \(M\cap\hat{M}=\varnothing\). \(Y = \{y_i\}_{i = 1}^{L}\) represents the label set. It should be noted that for general incomplete multimodal learning, all modalities are available and can be used during the training process. During testing, missing modalities \(\hat{X}_{\hat{M}}\) are unavailable as inputs, but they can serve as supervisory information to assist the training process of the model, thereby improving the performance of the model on incomplete modal data.
    
    To be consistent with previous research \cite{lian2023gcnet} for a fair comparison, on the multimodal dataset, we respectively adopt the pre-trained wav2vec \cite{schneider2019wav2vec}, pre-trained DeBERTa \cite{he2020deberta}, and pre-trained MA-Net \cite{zhao2021learning} as the feature extractors for the acoustic, lexical, and visual modalities. The subsequent elaboration of this paper is based on the three-modal dataset. Given a video clip, we input the acoustic (\(a\)), lexical (\(l\)), and visual (\(v\)) modalities into their respective feature extractors to obtain the samples \(X_m\) with the shape of \(B\times L\times D\), where \(m\in\{a, l, v\}\), \(B\) represents the batch size, \(L\) represents the sequence length, and \(D\) represents the feature dimension. 

    \textbf{Incomplete Multi-modal General Formulation.} Here, we summarize the general paradigm of existing incomplete multimodal learning in the field of MER. As shown in Fig. \ref{fig:settings} (a), such methods typically consist of the following components: a single-modality encoder \(E_m: X^M \to R^M\), which is responsible for learning the mapping from available modalities to their corresponding representations; a multimodal fusion module \(D_f: [R^{M_1}, R^{M_2}, R^{M_3}] \to R^{fusion}\); a modality-specific classifier or regressor \(C_m: R^M \to Y^M\), which maps the representation of the corresponding modality to the label; and a modality-fusion classifier \(C_f: R^{fusion} \to Y^{fusion}\), which maps the fused modality representation to the label. The optimization objective can be expressed as:
    \begin{align}\nonumber
    \min \; & l_{task}(C_f(D_f([R^{M_1},R^{M_2},R^{M_3}])), Y) \\\nonumber
    & + \sum_{m\in [M]}l_{task^m}(C_m(E(X^M)), Y)
    \end{align}
   Its function is to prompt the encoder \(E_m\) to extract joint representations that are consistent between the available and missing modalities. \(l_{task^m}\) is the task loss used to constrain the representation of specific modalities. However, within the same model, different modal combinations have different requirements for the characteristic information of unimodal representations, and these requirements may conflict with each other.

    \subsection{Overview Framework}
    To enable traditional multimodal learning architectures to effectively adapt to various modal combinations, we draw inspiration from the Low-Rank Adaptation (LoRA) technique widely used in language models \cite{hu2022lora}. This technique is typically employed to adapt models to new tasks. We construct an incomplete multimodal learning framework, which is specifically designed for efficient parameter adaptation in incomplete multimodal learning scenarios.
    
    As shown in Fig. \ref{fig:framework}, similar to existing incomplete multimodal learning methods \cite{xu2024leveraging,gao2024enhanced}, our model mainly consists of two components: multiple unimodal encoders and a cross-attention-based multimodal fusion module. In the MCLA module, we apply low-rank adaptation matrices to decouple the features extracted by the unimodal encoders. Subsequently, the characteristic unimodal feature information from specific modality combinations is fused with each other, and together with the shared unimodal information, assists the pre-trained model in completing the emotion recognition task. Meanwhile, to ensure that each unimodal representation is sufficiently learned across different modality combinations, in the DPFT module, we propose a dynamic adaptation scheme based on the difficulty of unimodal decoupling. During training, by dynamically adjusting the occurrence ratio of different modality combinations, the extraction of characteristic information in weaker modality combinations is promoted, enabling the model to achieve overall optimal performance within a limited training time.

\subsection{Low-Rank Adaptation for Incomplete Data}
\label{sec:preference}
    Low-Rank Adaptation (LoRA), proposed by Hu et al. \cite{hu2022lora}, is an efficient fine-tuning method. Its core is to freeze the weights of the pre-trained model during training and embed trainable low-rank decomposition matrices. Taking the linear layer \(\tilde{h}=Wh\) as an example, where \(W\in\mathbb{R}^{d_{out}\times d_{in}}\) is the pre-trained weight matrix, and \(d_{in}\) and \(d_{out}\) are the input and output dimensions respectively. LoRA adjusts the model parameters through low-rank matrices, and the process is as follows:
    \begin{equation}
    \tilde{h}=Wh+\Delta Wh = Wh+\alpha\cdot BAh
    \end{equation}
    $\Delta W = BA$ is the trainable weight adjustment matrix. Here, $A\in\mathbb{R}^{r\times d_{in}}$, $B\in\mathbb{R}^{d_{out}\times r}$, and $r\ll\min(d_{in},d_{out})$. $\alpha\geq1$ is used to regulate the impact of the weight adjustment matrix on the model output, and $BAh$ is the adaptive transformation of the input $x$.
    
    Based on this, we innovatively propose the \emph{modality combination aware low-rank adaptation} (MCLA) for unimodal data. This innovation aims to extract the characteristic information that unimodal data can provide in different modal combinations and, simultaneously, decouple the common information contained in different modal combinations.
    
      Specifically, we design multiple private adapters \(E^m_{prt}\) based on the number of modal combinations. For each modal combination, we locate the corresponding private adapter by inputting unimodal data and the modal combination index. Meanwhile, we employ a shared unimodal adapter \(E_{com}\). Regardless of the modal combination, the unimodal representation will be fed into this adapter to extract the common information of all modal combinations. The mathematical expressions are as follows. 
    \begin{align}
    R^m_{prt^i} &= E^m_{prt^i}(x^{m}) = A^m_{i}B^m_{i}x^{m}, \notag \\
    R^m_{com} &= E^m_{com}(x^{m}) = A^m_{com}B^m_{com}x^{m}
    \end{align}
    where \(i\) is the index of the currently input modal combination. \(A^m_{i}\) and \(B^m_{i}\) are the lower-projection and upper-projection matrices exclusive to the modal combination, and \(A^m_{com}\) and \(B^m_{com}\) are the lower-projection and upper-projection matrices shared by the modal combinations.
    
    However, the information of each modal combination may only be present in the corresponding \(R^m_{prt}\), which can render the commoninformation features of unimodalities meaningless. Given that the decoupled features respectively capture the features related to commonality or specific to modal combinations in unimodalities, we further formulate a soft orthogonality to reduce the information redundancy between homogeneous and heterogeneous features, while enhancing the learning of common information. Its expression is 
    \begin{equation}
    L_{ort} = \sum_{i\in [D]}\sum_{m\in [M]} \left [ \cos\left ( R^{m}_{com}, R^{m}_{prt^i} \right)- \cos\left ( R^{m}_{com}, R^{m} \right)\right]
    \end{equation}
   where, \([D]\) is the index set of the current modal combinations, \([M]\) is the total number of currently available modalities, and \(R^m\) is the unimodal representation result of the pre-trained model. 
    
    In the modal fusion stage, we separately integrate the common information and characteristic information of multiple modalities. The integrated representations are processed through a classifier, with the final prediction obtained by weighted summation of common prediction outputs and characteristic information features.
    \begin{align}
    \hat{y} &= F^{com}\big(\Omega\big([R^{m_a}_{prt}, R^{m_t}_{prt}, R^{m_v}_{prt}]\big)\big) \nonumber \\
    y_{com} &= F^{i}\big(\Omega\big([R_{com}^{m_a}, R_{com}^{m_t}, R_{com}^{m_v}]\big)\big)  \\
    y_{last} &= (1-weight)\times y_{com} + weight \times \hat{y} \nonumber
    \end{align}

    where $\hat{y}$ is the prediction result after the fusion of characteristic information, $y^{com}$ is the prediction result of common information, $[\cdots]$ represents the splicing operation. $F^{i}(\cdot)$ is the prediction head after fine-tuning adaptation for modal combinations, $F^{com}(\cdot)$ is the prediction head for common information, and $\Omega(\cdot)$ is responsible for fusing multimodal representations and outputting the fused features as prediction tokens. Here, $weight$ represents an adaptive parameter, which is derived by passing the characteristic representation \(R^{m}_{prt}\) through an MLP layer. It is used to balance the proportion of the common prediction result $y$ and the characteristic information prediction result $\hat{y}$ in the final prediction result $y_{last}$. 
    
    To optimize the above-mentioned fine-tuning modules, we achieve this by minimizing the loss of specific tasks. For example, in the emotion recognition task, the cross-entropy loss is used:
    \begin{equation}
    L_{task}=\text{CrossEntropy}(y, y_{last})
    \end{equation}
    While in the sentiment analysis task, the mean squared error (MSE) loss is used, denoted as:
    \begin{equation}
    L_{task}=\text{MSE}(y, y_{last})
    \end{equation}
    We integrate the above losses to achieve the optimization objective:
    \begin{equation}
    L_{total}=L_{task}+\beta \cdot L_{ort}
    \end{equation}
    where \(L_{task}\) is the task loss defined as the mean absolute error, and \(\beta\) controls the importance of different loss functions. The entire optimization process is implemented in an end-to-end manner. The specific training content for incomplete modal configurations can be found in the experimental section. 
    
    \subsection{Dynamic Parameter Fine-tuning}
    To train the above-mentioned low-rank adaptation module, the traditional active masking method usually provides training sets of the same scale for the adaptation of all modality combinations. Under training sets of the same scale, preferred combinations have more advantages in extracting characteristic information. Inspired by previous research on imbalanced multimodal learning, we evaluate the difficulty of decoupling unimodal data in modal combinations \cite{guo2024classifier,wei2024fly}, and then adjust the probabilities of different modal combinations appearing in the training set to balance the extraction of features of unimodal data under different modal combinations.
    
    Specifically, we quantify the degree of decoupling of unimodal data in combined representations using Jensen-Shannon divergence. When the similarity between the two is high, it means that the modal combination contains less discriminative characteristic information. Conversely, if the similarity is low, it indicates that the characteristic information of the modal combination has been effectively learned. The calculation process of the degree of decoupling is as follows: 
    \begin{equation}
    \begin{aligned}
    D_{KL}(p||m) &= \sum_{j} p_j \log\left(\frac{p_j}{m_j}\right)\\
    D_{KL}(q||m) &= \sum_{j} q_j \log\left(\frac{q_j}{m_j}\right)\\
    D_{JS}(p||q) &=  \left(D_{KL}(p||m) + D_{KL}(q||m)\right)\\
    s^i &=D_{JS}(R^m_{prt^i}||R^m_{com})
    \end{aligned}
    \end{equation}
     Here, \(m\) is the average distribution of the \(p\)-distribution and the \(q\)-distribution. Subsequently, the learning difficulty of each modality combination is quantified by measuring the differential changes in similarity between the common information of unimodalities and the characteristic information of different modality combinations during the training iteration process. The degree of decoupling of the unimodality in the test set is recorded as \(S_t=(s_{t}^{1},s_{t}^{ 2},s_{t}^{3},\cdots,s_{t}^{N})\), where \(i\) is the current iteration number and \(N\) is the total number of modal combinations. The calculation process of the difference between two consecutive \(s\) values is as follows: 
    \begin{equation}
    \begin{aligned}
    \Delta S_{t + 1} &= S_{t + 1}-S_{t} = (\Delta s_{t+1}^{1}, \Delta s_{t+1}^{2}, \cdots, \Delta s_{t+1}^{N}) \\
    &= (s_{t + 1}^{1}-s_{t}^{1}, \, s_{t + 1}^{2}-s_{t}^{2}, \, \cdots, \, s_{t + 1}^{N}-s_{t}^{N})
    \end{aligned}
    \end{equation}
    Here, \(t = 0, 1, 2,\cdots, T\), where \(T\) represents the total number of training iterations and the initial value is \(s_0 = 0\). By leveraging the learning difficulty of unimodal data within each modal combination, we dynamically adjust the occurrence probability of each modal combination in the training set during every training epoch. This adjustment is aimed at either enhancing or suppressing the learning intensity of unimodal data in each modal combination.
    
    For a model with \(m\) modalities, we rank all \(2^m\) modal combinations according to their learning difficulties based on the index. Then, we adaptively adjust the occurrence probability of each combination through a formula grounded on this ranking. 
    \begin {equation}
    \Delta q_{m}^{t + 1}=\begin {cases}-|q_{base}\cdot\lambda\cdot z (\Delta s_{t + 1})| & \text {if } idx>2^{m-1} \\+|q_{base}\cdot\lambda\cdot z (\Delta s_{t + 1})| & \text {if } idx<2^{m-1}\end {cases}
    \end {equation}
    Here, $q_{base}$ represents the basic dropout probability, which ranges between $(0, 1)$ and is used to control the range of the modal dropout probability, with different dropout probabilities corresponding to different rankings. $z(x)$ is a monotonically increasing function with a range between $(0, 1)$, and $\lambda>0$ determines the adjustment amplitude. For modal combinations with $idx > 2^{m-1}$, to prevent overfitting, we appropriately reduce their occurrence probability. Conversely, for modal combinations with $idx < 2^{m-1}$, we appropriately increase their occurrence probability to enhance the feature extraction ability of this modal combination. In the experiment, to avoid the occurrence probability of modalities becoming extremely small or extremely large, we dynamically set the maximum and minimum values of their occurrence probabilities as hyperparameters. We choose the Sigmoid(x) function as $z(x)$. 
    
	\section{Experiments}

\newcolumntype{C}[1]{>{\centering\arraybackslash}p{#1}}
\begin{table*}[htbp]
    \centering
    \centering
    \captionsetup{justification=justified, singlelinecheck=false, width=\textwidth} 
    \caption{Accuracy comparison under fixed missing protocol. "Average" shows mean performance across all modal combinations. Bold indicates best results; underlined values show second-best performance.}
   \begin{center}
   \setlength{\tabcolsep}{2.2pt}
   \begin{tabular}{c|c|c|c|c|c|c|c|c|c}
        \hline
        \multirow{3}{*}{Datasets} & \multirow{3}{*}{Models} & \multicolumn{8}{c}{Testing Condition} \\
        \cline{3-10}
        & \multicolumn{1}{c|}{} & $\{a\}$ & $\{t\}$ & $\{v\}$ & $\{a, v\}$ & $\{a, t\}$ & $\{t, v\}$ & Average & $\{a, t, v\}$ \\
        \cline{3-10}
         & \multicolumn{1}{c|}{} & ACC/F1 (\%) & ACC/F1 (\%)  & ACC/F1 (\%) & ACC/F1 (\%) & ACC/F1 (\%) &ACC/F1 (\%)  & ACC/F1 (\%) &ACC/F1 (\%)  \\
        \hline
        \multirow{7}{*}{CMU-MOSEI} 
            & MCTN\cite{pham2019found} & 62.70/54.50 & 82.60/82.80 & 62.60/57.10 & 63.70/62.70 & 83.50/83.30 & 83.20/83.20 & 73.05/70.60 & 84.20/84.20 \\
            & MMIN\cite{zhao2021missing} & 58.90/59.50 & 82.30/82.40 & 59.30/60.00 & 63.50/61.90 & 83.70/83.30 & 83.80/83.40 & 71.92/71.75 & 84.30/84.20 \\
            & GCNet\cite{lian2023gcnet} & 60.20/60.30 & 83.00/83.20 & 61.90/61.60 & 64.10/57.20 & 84.30/84.40 & 84.30/84.40 & 73.10/72.80 & 85.20/85.10 \\
            & IMDer\cite{wang2023incomplete} & 63.80/60.60 & 84.50/84.50 & 63.90/63.60 & 64.90/63.50 & 85.10/85.10 & 85.00/85.00 & 76.00/75.30 & 85.10/85.10 \\
            & DiCMoR\cite{wang2023distribution} & 62.94/60.43 & 84.26/84.35/ & 63.65/63.64 & 65.24/64.43 & 85.04/84.93 & 84.95/84.94 & 75.89/75.41 & 85.14/85.13 \\
            & MoMKE\cite{xu2024leveraging} &  \underline{65.92}/\underline{65.52} & 80.84/80.68 & 64.91/64.83 & 65.86/\underline{65.52} &  \underline{86.03}/\underline{85.91} &84.42/84.28 & 76.38/76.19 &  \underline{86.70}/\underline{86.59} \\
            & EUAR\cite{gao2024enhanced} & 64.54/60.73 & \underline{85.34}/\underline{85.22} &  \underline{66.35}/\underline{65.34} &  \underline{66.53}/65.42 & 85.62/85.14 &  \underline{86.02}/\underline{86.04} &  \underline{77.29}/\underline{76.33} & 86.62/86.43 \\
              & MCULoRA& \textbf{68.51/70.20} & \textbf{86.92/86.98} & \textbf{69.59/71.13} & \textbf{71.02/72.37}  & \textbf{87.19/87.26} & \textbf{87.02/87.14} & \textbf{79.63/80.34} & \textbf{87.19/87.26} \\ 
            & $\Delta_{SOTA}$ & {$\uparrow$2.59/$\uparrow$4.68} & {$\uparrow$1.58/$\uparrow$1.76} & {$\uparrow$3.24/$\uparrow$5.79} & {$\uparrow$4.49/$\uparrow$6.85} & {$\uparrow$1.16/$\uparrow$1.35} & {$\uparrow$1/$\uparrow$1.1} & {$\uparrow$2.34/$\uparrow$4.01} & {$\uparrow$0.49/$\uparrow$0.67} \\
        
    \hline
         & \multicolumn{1}{c|}{} & WA/UA (\%) & WA/UA (\%) & WA/UA (\%) & WA/UA (\%) & WA/UA (\%) & WA/UA (\%) & WA/UA (\%) & WA/UA (\%) \\
    \hline
        \multirow{7}{*}{IEMOCAP} 
            & MCTN\cite{pham2019found} & 49.75/51.62 & 62.42/63.78 & 48.92/45.73 & 56.34/55.84 & 68.34/69.46 & 67.84/68.34 & 58.94/59.13 & $-$ \\
            & MMIN\cite{zhao2021missing} & 56.58/59.00 & 66.57/ 68.02 & 52.52/\underline{51.60} & 63.99/65.43 & 72.94/75.14 & 72.67/73.61 & 64.10/65.24 &$-$ \\
            & IF-MMIM\cite{zuo2023exploiting} & 55.03/53.20 & \underline{67.02}/\underline{68.20} & 51.97/50.41 & 65.33/\underline{66.52} & \underline{74.05}/\underline{75.44} & 72.68/\underline{73.62} & 64.54/65.38 & $-$ \\
            & MRAN\cite{luo2023multimodal} & 55.44/57.01 & 65.31/66.42 & 53.23/49.80 & 64.70/64.46 & 73.00/74.58 & 72.11/72.24 & 63.97/64.08 & $-$ \\
            & MoMKE\cite{xu2024leveraging} & \underline{56.97}/\underline{60.42} & 64.54/66.05 & \underline{53.33}/47.62 & \underline{66.98}/64.69 & 72.38/74.35 & \underline{73.17}/71.79 & \underline{66.51}/\underline{66.10} & \underline{78.20}/\underline{77.78} \\
              &MCULoRA&  \textbf{67.30/68.66} & \textbf{75.27/75.75} & \textbf{55.68/52.52} & \textbf{71.94/72.06} & \textbf{79.96/81.29} & \textbf{76.98/77.71} & \textbf{72.55/72.85} & \textbf{80.75/81.96} \\
            & $\Delta_{SOTA}$ &$\uparrow$10.33/$\uparrow$8.24& $\uparrow$ 8.25/$\uparrow$7.55 & $\uparrow$2.35/$\uparrow$0.92 & $\uparrow$4.96/$\uparrow$5.54 & $\uparrow$5.91/$\uparrow$5.85 & $\uparrow$3.79/$\uparrow$4.09 & $\uparrow$6.04/$\uparrow$6.75 & $\uparrow$2.55/$\uparrow$4.18 \\
        \hline
    \end{tabular}
\end{center}
\end{table*}

\subsection{Datasets and Evaluation Metrics}
    In this section, we evaluate the performance of our proposed model \textsf{MCUloRA} and seven state-of-the-art methods.
    All methods were implemented in Python. The experiments were conducted in an Intel Core 2.90GHz server with A40 48GB (GPU) and 256GB RAM, running Ubuntu 18.04 system.
    
    To evaluate the effectiveness of the novel framework MCULoRA, we conduct experiments on three benchmark multimodal emotion recognition or sentiment analysis datasets, namely IEMOCAP and CMU-MOSEI:
    \begin{enumerate}
    \item \textbf{IEMOCAP} \cite{menze2014multimodal}: The IEMOCAP dataset comprises five dyadic dialogue sessions, during which actors engage in improvisational performances or act in accordance with pre-scripted scenarios. Each dialogue is subsequently segmented into a multitude of utterances, and each of these utterances is meticulously annotated with categorical emotion labels. In order to ensure a fair comparative analysis, we adhere to the methodologies adopted in previous research works [12, 36, 39] to establish an experimental configuration for the recognition of four distinct emotion categories, namely happiness, sadness, neutrality, and anger.
    \item \textbf{CMU-MOSEI} \cite{zhuang2018multivariate,qiu2023myops}: The CMU-MOSEI dataset  encompasses 22,856 video clips contributed by over 1,000 online YouTube content creators. Out of these, 16,326 clips are designated for the training phase, while the remaining 1,871 and 4,659 clips are respectively employed for validation and testing purposes. All the utterances within this dataset are randomly selected from monologue videos covering a diverse range of topics and follow the identical annotation scheme as that of the CMU-MOSI dataset, where the sentiment scores are delineated within the range of [-3, 3].
    \end{enumerate}
	
        In terms of the selection of evaluation metrics, for the IEMOCAP dataset, we use the weighted accuracy (WA) and unweighted accuracy (UA) for measurement.
        For the CMU-MOSEI dataset, referring to previous studies \cite{lian2023gcnet,luo2023multimodal,wang2023incomplete}, we focus on the positive-negative emotion classification task, classifying emotion scores less than 0 as negative and those greater than 0 as positive, and use accuracy (ACC) and F1-score as evaluation metrics.
    
     Following the literature \cite{lian2023gcnet,pham2019found,zhao2021missing}, we investigate the performance of different methods under two commonly used protocols: (1) Fixed Missing Protocol; (2) Random Missing Protocol. Under the Fixed Missing Protocol, we consistently discard either one modality (i.e., {text (l), vision (v)}, {text (l), audio (a)}, {vision (v), audio (a)}) or two modalities (i.e., {text (l)}, {vision (v)}, {audio (a)}). For the Random Missing Protocol, the missing pattern for each sample is randomly determined (i.e., each sample may be missing one or two modalities), and the seed is set to 66.The optimal setting for \(\beta\) is set to 0.001 via the performance on the validation set. 
	
	\subsection{Implementation Details}
    For both the IEMOCAP and CMU-MOSEI datasets, the dimension \(d\) of their modality representations is set to 256. Specifically, the modality masking probability for the IEMOCAP dataset ranges from 0.45 to 0.55, while for the CMU-MOSEI dataset, it ranges from 0.4 to 0.6.
    During the pre-training phase, when the batch size for the IEMOCAP dataset is set to 16 and that for the CMU-MOSEI dataset is set to 32, the maximum number of training epochs for the IEMOCAP dataset is set to 300, and for the CMU-MOSEI dataset, it is set to 200. The maximum number of training epochs in the fine-tuning phase is the same as that in the pre-training phase. In the basic settings, the rank of fine-tuning is set to 4. The number of blocks \(Q\) in the Transformer and the number of heads in the multi-head self-attention mechanism are set to 4 and 2 respectively.
    Regarding the optimizer, we choose the Adam optimizer and set the learning rate to 0.0001. For all datasets, the dropout rate is uniformly set to 0.5. For the IEMOCAP dataset, we conduct five-fold cross-validation using the leave-one-session-out strategy. For the CMU-MOSEI dataset, each experiment is repeated five times, and the average results are reported. 
	
\subsection{Comparison Study}

To comprehensively and systematically evaluate the performance of MCULoRA under various incomplete multimodal conditions, we conducted an in-depth comparative analysis based on two benchmark datasets. We compared MCULoRA with the current state-of-the-art (SOTA) methods that follow the fixed-missing protocol and the random-missing protocol. The methods involved in this comparative experiment include MCTN \cite{pham2019found}, MMIN \cite{zhao2021missing}, IF-MMIN \cite{zuo2023exploiting}, MRAN \cite{luo2023multimodal}, GCNet \cite{lian2023gcnet}, IMDer \cite{wang2023incomplete}, DiCMoR \cite{wang2023distribution}, MoMKE \cite{xu2024leveraging}, and EUAR \cite{gao2024enhanced}. It should be specifically noted that, in the reproduction of the compared algorithms, MoMKE used a dataset with complete modalities during the training process. The experimental content of the random-missing protocol is presented in the supplementary materials. 

Regardless of whether it is under the fixed-missing protocol or the random-missing protocol, MCULoRA has demonstrated leading and outstanding performance in almost all evaluation metrics, strongly confirming its superiority.

As can be seen from the experimental results of the fixed-missing protocol shown in Table 1, the average improvement of the weighted accuracy (WA)/accuracy (ACC) of MCULoRA on the two datasets reached 2.34\% and 6.04\% respectively. Under relatively difficult-to-learn incomplete multimodal conditions such as {audio, video}, {audio}, and {video}, MCULoRA still has performance advantages over other methods. This is because MCULoRA has the ability to assess the difficulty of decoupling the information of a single modality in each modal combination and can dynamically enhance the adaptation of weak modal combinations, thus significantly improving the performance of the corresponding combinations.

	\begin{table*}[htbp]
    \centering
    \captionsetup{justification=centering, singlelinecheck=false, width=\textwidth} 
    \caption{Ablation results of MCLA module and DPFT Strategy.}
    \vspace{-0.05in}
    \setlength{\tabcolsep}{2.3pt}
   \begin{center}
\begin{tabular}{c|c c|c|c|c|c|c|c|c|c}
        \hline
        \multirow{3}{*}{Datasets} & \multirow{3}{*}{MCLA}& \multirow{3}{*}{DPFT} & \multicolumn{8}{c}{Testing Condition} \\
        \cline{4-11}
        & &  & $\{a\}$ & $\{t\}$ & $\{v\}$ & $\{a, v\}$ & $\{a, t\}$ & $\{t, v\}$ & Average & $\{a, t, v\}$ \\
        \cline{4-11}
         & &  & ACC/F1 (\%)  & ACC/F1 (\%)  & ACC/F1 (\%)  &ACC/F1 (\%) & ACC/F1 (\%) & ACC/F1 (\%)  & ACC/F1 (\%)  &ACC/F1 (\%) \\
         
        \hline
         \multirow{4}{*}{CMU-MOSEI} 
            &-& -& 62.80/64.33 & 80.85/80.87 & 56.86/57.40 & 62.23/63.34 & 86.60/86.59 & 79.41/78.90 & 73.42/73.80& 85.17/85.20 \\
            &\checkmark&  -& 67.04/69.29 & 86.07/86.13 & 66.23/67.41 & 68.99/69.92 & 862.9/86.40 & 85.89/85.99 & 78.18/78.85 & 86.71/86.81 \\
            &-&\checkmark& 67.32/69.48 & 85.57/85.44 & 65.12/68.79 & 67.88/70.88 & 85.99/85.91 &86.52/86.51 & 77.87/79.10 & 86.70/86.73 \\
           &\checkmark& \checkmark& \textbf{68.51/70.20} & \textbf{86.92/86.98} & \textbf{69.59/71.13} & \textbf{71.02/72.37}  & \textbf{87.19/87.26} & \textbf{87.02/87.14} & \textbf{79.63/80.34} & \textbf{87.19/87.26} \\
        \hline
                
         & & \multicolumn{1}{c|}{} & WA/UA (\%) & WA/UA (\%) & WA/UA (\%) & WA/UA (\%) & WA/UA (\%) & WA/UA (\%) & WA/UA (\%) & WA/UA (\%) \\
    \hline
        \multirow{4}{*}{IEMOCAP} 
            & -&-& 45.74/48.38 & 70.44/71.62 &51.12/47.48 & 62.51/61.14 & 76.22/77.32 & 75.40/75.53 & 65.89/65.93 & 79.76/80.06 \\
            & \checkmark& -& 55.41/58.38 & 73.61/73.88 & 53.54/50.49 & 68.77/68.23 & 79.44/80.36 & 76.94/76.54 & 69.68/69.90 & 80.05/81.42 \\
            &-& \checkmark& 58.72/61.71 & 74.56/74.40 & 52.28/49.77 & 66.62/65.24 & 78.89/79.33 & 75.73/74.46 & 69.43/69.09 & 79.19/78.69 \\
            &\checkmark& \checkmark& \textbf{67.30/68.66} & \textbf{75.27/75.75} & \textbf{55.68/52.52} & \textbf{71.94/72.06} & \textbf{79.96/81.29} & \textbf{76.98/77.71} & \textbf{72.55/72.85} & \textbf{80.75/81.96} \\
        \hline
       
   \end{tabular}
   \end{center}
\end{table*}
	\subsection{Ablation Study}

        \begin{figure}[!t] 
            \centering
            \includegraphics[width=0.85\columnwidth]{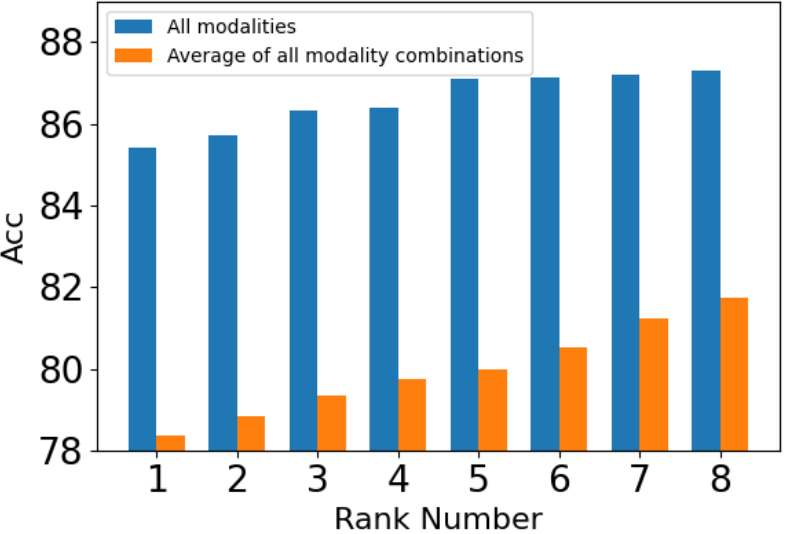}
            \vspace{-0.05in}
            \caption{Ablation study on the rank number for adapter fine-tuning. We conducted experiments on the CMU-MOSEI dataset and tested the emotion recognition accuracy with the rank number of the feature adapter fine-tuning matrix ranging from 1 to 8.}
            \label{fig:rank}
            \vspace{-0.1in}
    \end{figure}
    
    In this subsection, we conduct ablation studies to deeply explore the roles played by various configurations and expert modules in MCULoRA, aiming to reveal the relationship between its internal mechanisms and performance.
    
    \textbf{Ablation of the MCLA and DPFT:} To thoroughly analyze the effectiveness of two key designs in MCULoRA, \emph{modality combination aware low-rank adaptation} (MCLA) module  and \emph{dynamic parameter fine-tuning} (DPFT) Strategy, we carefully plan and carry out the following ablation experiments:
    \begin{enumerate}[topsep=5pt, partopsep=0pt]
   \item  Without MCLA training: Under this setting, unimodalities no longer extract the characteristic information corresponding to modal combinations and do not extract the common information of each modal combination for auxiliary prediction. Only the pre-trained encoder module is enabled to process the available modalities.
   \item  Without DPFT training: The model structure used in this experiment is the same as that of MCULoRA, but the probability of each modal combination appearing in the training set is not adjusted. That is, the probabilities of all modalities appearing remain the same, and the training epoch settings are also kept consistent.
    \end{enumerate}

        \begin{figure}[!t] 
            \centering
            \includegraphics[width=1\columnwidth]{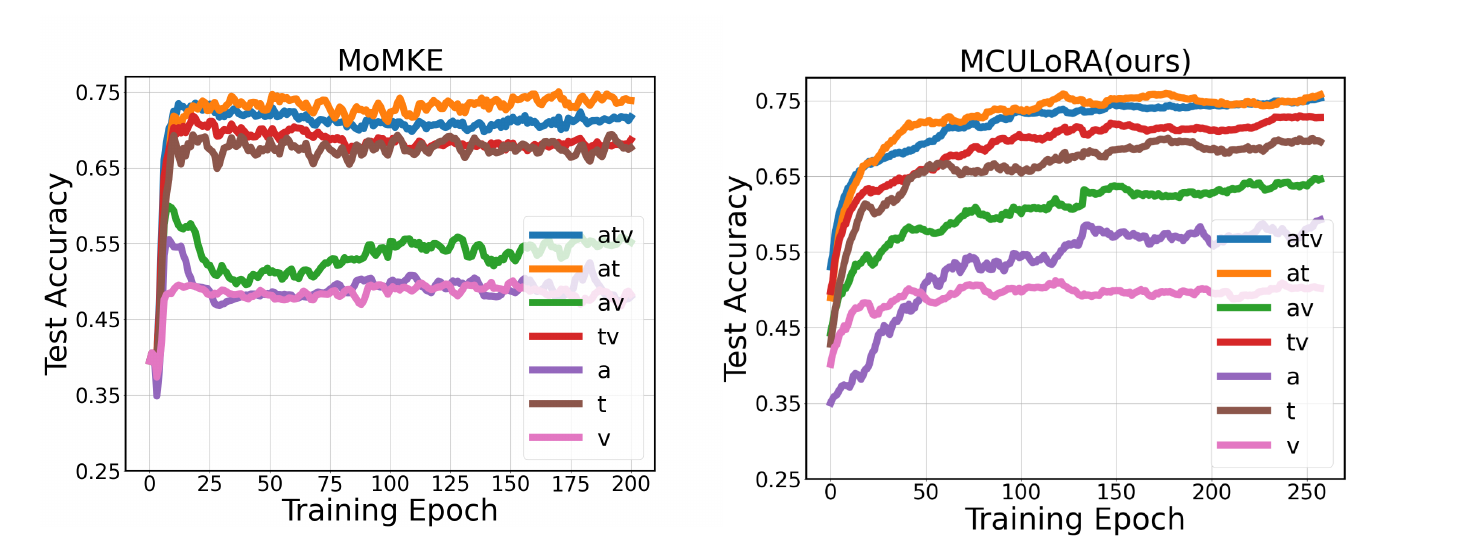} 
            \caption{Analysis of the training convergence. The performance fluctuations of different models in various modal combinations during the training process are observed. }
            \label{fig:train}
    \end{figure}
    The results of the ablation experiments are shown in Table 3. During the training process, when the MCLA module based on unimodalities is removed, a significant performance decline can be clearly observed. This indicates that multimodal training causes unimodalities to lose their characteristic information, resulting in a lack of discriminative information in the representations of some modality combinations and a substantial reduction in prediction accuracy. Supplementing the characteristic information of unimodalities can help the model learn more comprehensive representations, thus effectively improving the overall performance of MCULoRA.
    
    Compared with the above two ablation scenarios, when the DPFT strategy is not adopted, the performance decline of the model is more pronounced. This means that the difficulty of characteristic adaptation of unimodalities varies in each modality combination situation. When this strategy is not applied, the characteristic adaptation of each modality combination is different, leading to poor extraction of characteristic information for some modality combinations. The Dynamic Low-Rank Adaptation (LoRA) strategy can dynamically adjust the probability of modality combinations appearing in the training set according to the characteristics of incomplete scenarios, strengthen the extraction of characteristic information of weak modality combinations, and thus significantly enhance the multimodal representation effect. 
    
    \textbf{Ablation of the adaptation matrix rank number:} To explore the impact of the rank number of the adaptation matrix for each unimodal encoder on modal representations, we conducted experiments using adaptation matrices with different rank numbers during the training period. Fig. \ref{fig:rank} shows the average experimental results under various modality-missing conditions and the results when all modalities are available on the MOSEI dataset. As the rank number of the adaptation matrix increases, the improvement is limited when all modalities are present, but the prediction accuracy for modality-missing conditions increases significantly. The main reason is that the increase in the rank number enables unimodalities to extract more characteristic information, greatly enriching the discriminative information of weak modal combinations. This result strongly validates our hypothesis that the characteristic information provided by unimodal representations varies in different modal combinations.

    \textbf{Training Process Analysis:} As shown in Fig. \ref{fig:train}, we also investigated the performance fluctuations of the proposed model during the training process on the IEMOCAP dataset. The results indicate that, in contrast to other state-of-the-art (Sota) models, the training of our complete model exhibits a smoother upward progression and ultimately achieves significantly better performance. Specifically, the training processes of the two comparison models show that, compared with MoMKE, our training for modal combination adaptation has better performance under all modal combination conditions. The reason is that during the mixed training stage, MoMKE cannot effectively handle the conflicts of characteristic information between multimodal mixed training and unimodal training. This leads to a slow increase or even a decrease in the accuracy of weaker unimodalities (such as speech and video) during the training process. In contrast, our model decouples the unimodal characteristic information, enabling the performance of all modal combinations to steadily improve during the training process. 
    
    In conclusion, the above-mentioned analytical experiments once again strongly demonstrate the rationality and effectiveness of the proposed model.
	\subsection{Case Study}
	\begin{figure}[!t] 
            \centering
            \setlength{\belowcaptionskip}{5pt}
            \includegraphics[width=0.92 \columnwidth]{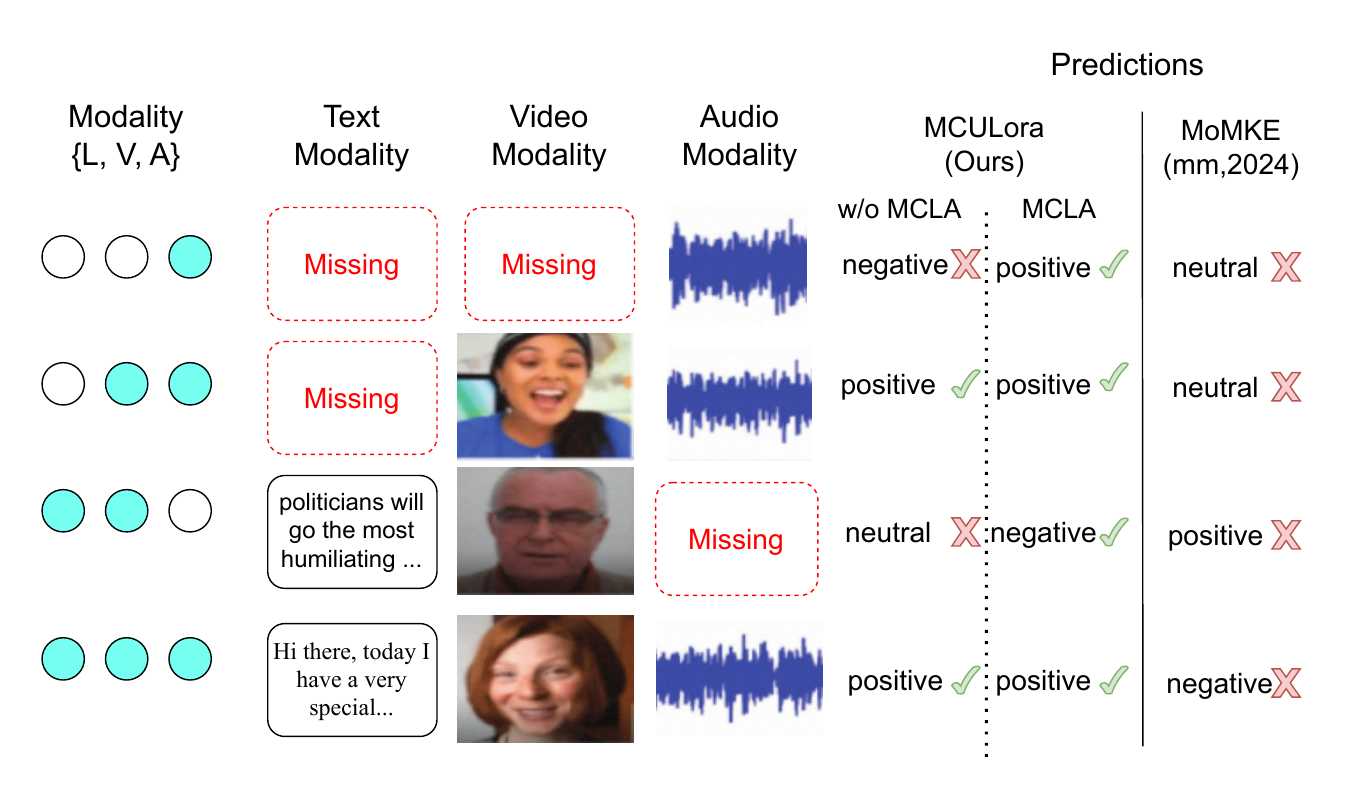} 
            \caption{Visualization of the test cases selected from the CMU-MOSEI dataset. It can be observed that our MCULora method, after the supplementation of characteristic information, can make more effective predictions. }
            \label{fig:case}
    \end{figure}
    To further clarify the necessity of using unimodal characteristic information to assist in emotion recognition, we present several typical test cases from the CMU-MOSEI dataset with different modal-missing situations. It should be specifically noted that when a certain modality is unavailable in the combination, we mark it with a red dashed-line rectangle. As shown in Fig. \ref{fig:case}, we compare the proposed model with the recent similar model MoMKE \cite{xu2024leveraging}. Under various modal-missing scenarios, our MCULoRA method can always achieve accurate classification, while MoMKE has difficulty in accurately completing the task. Notably, with the assistance of the unimodal characteristic information provided by the MCLA module, the accuracy of the model has been further improved. This indicates that our model can extract more discriminative information from multimodal data with diverse modal combinations.
	
	\section{Conclusion}
	
    In this paper, we propose a novel incomplete multimodal learning framework named \emph{unimodal decoupled dynamic low-rank adaptation} (MCULoRA). This framework encompasses a modality combination aware low-rank adaptation module and a shared low-rank adaptation module for unimodalities. Its aim is to efficiently decouple the characteristic information of unimodalities in different combinations within the MER task, thus enabling multimodal pre-trained models to achieve accurate predictions. Meanwhile, to alleviate the imbalance problem of unimodalities during the adaptation process of modality combinations, we assess the difficulty of decoupling unimodalities in different modality combinations and dynamically adjust the overall appearance probability of modality combinations during training accordingly. Experiments have fully verified that MCULoRA exhibits remarkable robustness in various incomplete multimodal scenarios. Ablation experiments and visualization analyses indicate that its superiority stems from the effective supplementation of unimodal characteristic information into the joint representation. However, currently, during the low-rank adaptation process, MCULoRA still assumes that all modality data are available. In practical applications, it is rather difficult to collect complete modality training data. Future work will address low-rank adaptation with inherently incomplete multimodal training data.
    \clearpage

    \bibliographystyle{ACM-Reference-Format}
    \bibliography{references}

\end{document}